%% file: main.tex
\documentclass{article}

\usepackage[preprint]{neurips_2025}
\usepackage[utf8]{inputenc} % allow utf-8 input
\usepackage[T1]{fontenc}    % use 8-bit T1 fonts
\usepackage{url}          
\usepackage{hyperref}
\usepackage{amsmath,amssymb,amsfonts}
\usepackage{algorithmicx}
\usepackage{algorithm}
\usepackage{algpseudocode}
\usepackage{graphicx}
\usepackage{textcomp}
\usepackage{xcolor}
\usepackage{multirow}
\usepackage{pifont}
\newcommand{\cmark}{\ding{51}} % ✓
\newcommand{\xmark}{\ding{55}} % ✗
\usepackage{bm}
\usepackage{booktabs}
\usepackage{placeins}

\renewcommand{\cite}{\citep}

\title{Accelerating Long-Term Molecular Dynamics with Physics-Informed Time-Series Forecasting}

\author{
  Hung Le\thanks{Corresponding author: \texttt{thai.le@deakin.edu.au}} \\
  Applied Artificial Intelligence Initiative \\
  Deakin University \\
  Geelong, Australia \\
  \And
  Sherif Abbas \\
  Applied Artificial Intelligence Initiative \\
  Deakin University \\
  Geelong, Australia \\
  \And
  Minh Hoang Nguyen \\
  Applied Artificial Intelligence Initiative \\
  Deakin University \\
  Geelong, Australia \\
  \And
  Van Dai Do \\
  Applied Artificial Intelligence Initiative \\
  Deakin University \\
  Geelong, Australia \\
  \And
  Huu Hiep Nguyen \\
  Applied Artificial Intelligence Initiative \\
  Deakin University \\
  Geelong, Australia \\
  \And
  Dung Nguyen \\
  Applied Artificial Intelligence Initiative \\
  Deakin University \\
  Geelong, Australia \\
}
\begin{document}

\maketitle

\begin{abstract}
\input{p_abs}

\end{abstract}

\section{Introduction}
\input{p_intro}

\section{Method}
\input{p_method}

\section{Experimental Results}
\input{p_exp}

\section{Related Works}
\input{p_related}

\section{Discussion}
\input{p_discuss}

\section{Acknowledgment}
This research was funded (partially or fully) by the Australian Government through the Australian Research Council.
Dr Hung Le is the recipient of an Australian Research Council Discovery Early Career Researcher Award (project number DE250100355) funded by the Australian Government.
\begin{figure*}[t]
    \centering
    \includegraphics[width=0.8\linewidth, trim=0 240 0 0, clip]{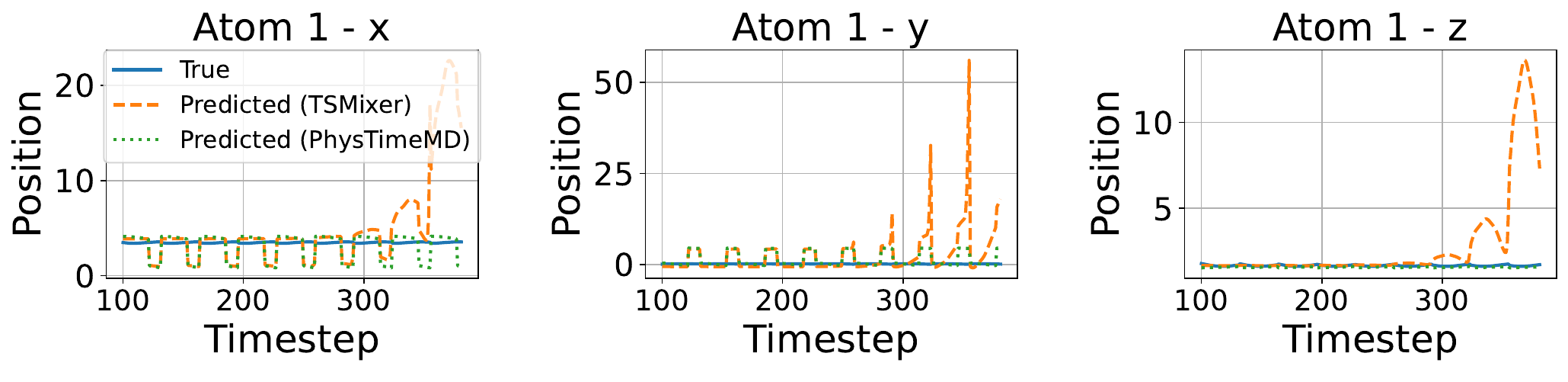}
    \caption{Comparison of TSMixer and PhysTimeMD-enhanced TSMixer on Dataset E. We visualize the predicted positions of an atom (the same observation was observed for other atoms), each along the three Cartesian axes. The model TSMixer without physics-informed regularization ($\lambda=0$) exhibits divergence (after 200 timesteps), while incorporating the PhysTimeMD term stabilizes rollout and significantly reduces error.}
    \label{fig:appendix_fail_case_with_update}
\end{figure*}

\bibliographystyle{plainnat}
\bibliography{sample}

% \clearpage
\section{Appendix}
\input{p_appendix}

\end{document}

%% file: p_abs.tex
Efficient molecular dynamics (MD) simulation is vital for understanding atomic-scale processes in materials science and biophysics. Traditional density functional theory (DFT) methods are computationally expensive, which limits the feasibility of long-term simulations. We propose a novel approach that formulates MD simulation as a time-series forecasting problem, enabling advanced forecasting models to predict atomic trajectories via displacements rather than absolute positions. We incorporate a physics-informed loss and inference mechanism based on DFT-parametrised pair-wise Morse potential functions that penalize unphysical atomic proximity to enforce physical plausibility. Our method consistently surpasses standard baselines in simulation accuracy across diverse materials. The results highlight the importance of incorporating physics knowledge to enhance the reliability and precision of atomic trajectory forecasting. Remarkably, it enables stable modeling of thousands of MD steps in minutes, offering a scalable alternative to costly DFT simulations.

%% file: p_intro.tex
Molecular dynamics (MD) simulations are critical for examining atomic-scale interactions in many scientific fields such as materials science and drug discovery. Classical MD, based on numerically integrating Newton’s equations of motion at the atomic scale, has long been a fundamental tool for probing molecular behavior \cite{verlet1967computer,mccammon1977dynamics}. Despite its generality, this method offers only an approximate description of how atoms interact with one another. To be stable, it needs very small time steps, on the order of femtoseconds. This results in high computational costs when simulating longer-timescale phenomena. On the other hand, ab initio molecular dynamics (AIMD) methods such as density functional theory (DFT) provide a more accurate quantum-level description of atomic interactions but are orders of magnitude more expensive, making them impractical for large systems or long simulations \cite{marx2000ab, kuhne2014second}. Both approaches, therefore, face fundamental limitations in scalability and efficiency. As a result, the growing demand for faster and more scalable MD modeling has led to increasing interest in machine learning (ML)-based alternatives \cite{wang2020accelerated,audagnotto2022machine}.
State-of-the-art ML methods, including generative models \cite{endo2018multi,jing2024generative}, graph neural networks \cite{zheng2021learning,maji2025accelerating}, and recurrent architectures \cite{eslamibidgoli2019recurrent,tsai2020learning,wang2020accelerated}, have demonstrated potential in deciphering molecular interactions from existing MD trajectory data. \textit{Nonetheless, three limitations remain}. First, most current sequential MD models predict exact atomic positions, which makes it hard for them to work with different molecule structures and typically leads to unstable long-term rollouts. This challenge arises from position-based models relying heavily on the spatial patterns seen during training, making them particularly sensitive to distribution shifts at test time. As prediction errors accumulate over successive steps, these models can quickly diverge from reasonable trajectories. Second, generative approaches suffer from exposure bias and dependence on carefully tuned priors~\cite{endo2018multi, nam2024flow}, which hinder their robustness and scalability. Third, all existing approaches lack explicit physical constraints, leading to unrealistic atomic configurations that fail to align with real-world molecular behavior. Without these constraints, models may generate unstable high-energy states, where atoms exhibit unphysical overlaps or diverge from expected motion patterns.

To overcome these limitations, we propose a physics-informed time-series forecasting framework, dubbed \textbf{PhysTimeMD}, that treats molecular dynamics as a time-series prediction problem. Moreover, by forecasting atomic displacements instead of coordinate positions, our approach captures relative motion and encodes local physical dynamics while remaining invariant to global spatial shifts. Rather than sampling atomic trajectories, our approach formulates molecular dynamics as a deterministic displacement forecasting task. This avoids handcrafted prior tuning and complicated sampling processes in generative rollouts, leverages local temporal dependencies, and enables the incorporation of physical constraints later. By reframing the problem in this way, we improve simulation robustness across temperatures and molecular systems while benefiting from advances in time-series forecasting~\cite{le2018dual, leneural, zeng2023transformers,wutimesnet,wang2024timemixer}, offering a stable and scalable framework for long-range trajectory prediction. More importantly, we incorporate explicit physical constraints by integrating knowledge from the Morse potential  \cite{morse1929diatomic} into both the training loss and the inference process. Specifically, the loss function penalizes atomic pairs that violate realistic bonding distances, while the inference mechanism adjusts predicted displacements to discourage unphysical overlaps and excessively high interatomic forces. This dual enforcement of physical plausibility not only reduces the likelihood of unstable or energetically unfavorable configurations but also promotes more accurate and chemically meaningful trajectory generation throughout extended simulations. By combining state-of-the-art time-series forecasting with physics-informed regularization, our method outperforms existing ML approaches that model atom positions without physical insights, delivering more accurate, physically consistent, and computationally efficient molecular dynamics simulations.

In summary, our contributions are three-fold:
\begin{itemize}
    \item We formulate MD simulation as a displacement-based time-series forecasting framework, allowing application of state-of-the-art time-series models while avoiding the complexity and instability associated with generative sampling methods.
    \item We pioneer incorporating physics-informed constraints by integrating the Morse potential knowledge into both the training loss and inference, promoting realistic atomic interactions.
    \item We conduct extensive experiments across diverse materials, showing that our method achieves higher accuracy and can generate stable molecular trajectories for thousands of steps, significantly outperforming existing ML-based MD approaches in both precision and efficiency.

\end{itemize}

%% file: p_method.tex
\begin{figure*}[t]
    \centering
    \includegraphics[width=1\linewidth]{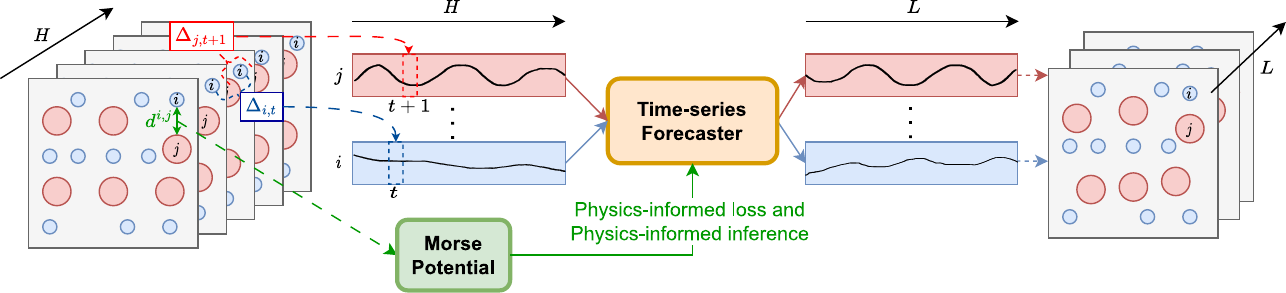}
    \caption{Overview of the PhysTimeMD framework. The framework takes as input a sequence of atomic positions over the past $H$ timesteps and converts them into relative displacements for each atom (e.g., $\Delta_{i,t}$ and $\Delta_{j,t+1}$ represent the movement vectors of atoms $i$ and $j$ between consecutive steps). Then, it uses a time-series forecaster to predict $L$-step future displacements. These predictions are then integrated to reconstruct future atomic positions (see Sec. \ref{sec:framework}). A physics-informed mechanism, guided by the Morse potential based on the distance between 2 atoms (e.g, $d^{i,j}$), refines the predicted configurations through a physics-based loss during training (see Sec. \ref{sec:pit}) and a physics-based correction during inference (see Sec. \ref{sec:pii}), ensuring adherence to physical constraints.
}
    \label{fig:phys_time_md}
\end{figure*}

We introduce the PhysTimeMD framework, which predicts future atomic positions by combining time-series learning and physics-informed regularization. It processes past atomic positions into displacement vectors, forecasts future movements with any time-series models, and reconstructs future positions. A final correction step, based on the Morse potential, ensures the predicted configurations obey physical interaction constraints during both training and inference. An overview of our framework is depicted in Fig. \ref{fig:phys_time_md}

\subsection{Displacement-based Time-series Forecasting Formulation}\label{sec:framework}

Let $\mathcal{S} = \{ S_t \}_{t=1}^{T}$ denote a molecular trajectory, where each state $S_t$ consists of $N$ atoms with atomic positions $\mathbf{r}_t = \{ \mathbf{r}_t^i \}_{i=1}^{N}$ in three-dimensional space, i.e, $\mathbf{r}^i_t=\{P^x_{i,t},P^y_{i,t},P^z_{i,t}\}$. The goal of forecasting is to predict future states $S_{t+1}, \dots, S_{T}$ given historical observations $S_1, \dots, S_t$. In practice, to enable efficient training, fixed-size input and output windows are typically used. A generic forecaster $f_\theta$ models the conditional distribution of future states given past observations:
\begin{equation}\label{eq:fcp}
\hat{S}_{t+1:t+L} = f_\theta(S_{t-H:t}),
\end{equation}
where $H$ is the historical window size, $L$ is the forecasting horizon, and $f_\theta$ is a parameterized function that captures temporal patterns within the data. However, we argue that predicting absolute atomic positions is not robust due to the sensitivity to global transformations and the accumulation of errors over time. Instead of directly predicting absolute positions $\mathbf{r}_{t+1}$, we model atomic displacements $\Delta_t$ as:
\begin{equation}
\Delta_t^i = \mathbf{r}_{t+1}^i - \mathbf{r}_t^i = 
\begin{bmatrix}
P_{i,t+1}^{x} - P_{i,t}^{x} \\
P_{i,t+1}^{y} - P_{i,t}^{y} \\
P_{i,t+1}^{z} - P_{i,t}^{z}
\end{bmatrix}
= 
\begin{bmatrix}
\Delta_{i,t}^{x} \\
\Delta_{i,t}^{y} \\
\Delta_{i,t}^{z}
\end{bmatrix}, 
\quad i = 1,.., N.
\end{equation}
This displacement-based formulation improves generalization by ensuring predictions remain invariant to global coordinate transformations and enhances stability by reducing the accumulation of positional errors.

To construct the training dataset, we process molecular trajectory data as follows. Given $T$ molecular configurations $\{ S_t \}_{t=1}^{T}$, we extract both absolute positions and displacements. Each sample consists of the atomic coordinates $\mathbf{r}_t$ at time step $t$ and the corresponding displacements $\Delta_t$. The dataset is structured as a time series with features at step $t$ as:
\begin{equation}
\begin{aligned}
X_t &= [\mathbf{r}_t, \Delta_{t-1}]\\
    &= \{P_{i,t}^x, P_{i,t}^y, P_{i,t}^z, \Delta_{i,t-1}^x, \Delta_{i,t-1}^y, \Delta_{i,t-1}^z \}_{i=1}^N   
\end{aligned}
\end{equation}
where $P$ and $\Delta$ denote position and displacement components, respectively. At $t=1$, we set $\Delta_{0}=\mathbf{0}$. The target becomes:
\begin{equation}
\Delta_{t:t+L-1} = \{\Delta_{i,t:t+L-1}^x, \Delta_{i,t:t+L-1}^y, \Delta_{i,t:t+L-1}^z \}_{i=1}^N,
\end{equation}
which fully determines $S_t$ since the original states can be reconstructed from the predicted displacements and the original position $\{P_{i,1}^x, P_{i,1}^y, P_{i,1}^z \}_{i=1}^N$.

We translate the forecasting problem (Eq. \ref{eq:fcp}) into predicting the displacement target series $\Delta_{t:t+L-1}$ given the input $X_{t-H:t}$. Specifically, we train a model $f_\theta$ to learn the mapping: 
\begin{equation} 
\hat{\Delta}_{t:t+L-1} = f_\theta(X_{t-H:t}).
\end{equation} 

By learning displacement dynamics, the model captures local interactions while remaining invariant to rigid-body transformations. The predicted states $\hat{S}_{t+1:t+L}$ can be reconstructed from the predicted displacements and the initial positions:
\begin{equation}\label{eq:S_recon}
\hat{S}_{t+1} = \{ \hat{\mathbf{r}}^i_{t+1} \}_{i=1}^{N} = \left\{ \mathbf{r}^i_{1} + \sum_{s=1}^{t} \hat{\Delta}_{i,s} \right\}_{i=1}^{N}.
\end{equation}

During inference, we adopt an autoregressive strategy over fixed forecasting windows of length $L$. Specifically, at each autoregression step, the model takes as input the recent window of atomic positions and displacements, predicts the displacements for the next interval, and reconstructs future atomic coordinates via Eq. \ref{eq:S_recon}. These predicted positions, combined with their corresponding displacements, are used as features for the next prediction window. This iterative procedure enables long-range trajectory rollout to arbitrary trajectory lengths. By forecasting displacements segment by segment rather than sampling full trajectories in one pass, our approach avoids compounding generation noise and retains the flexibility to incorporate physical constraints or corrections between steps.

\subsection{Physics-Informed Training (PIT)}\label{sec:pit}

\textbf{Morse Potential}
To ensure physically consistent predictions, we introduce a physics-informed regularization term derived from the Morse potential, which models interatomic interactions as:
\begin{equation}\label{eq:morse}
E(d) = D_e \left(1 - e^{-a(d - d_e)}\right)^2 + b,
\end{equation}
where $d$ is the interatomic distance, $D_e$ is the bond dissociation energy, $d_e$ is the equilibrium bond length, $a$ controls the steepness of the potential, and $b$ is a constant energy offset. We fit these parameters ($d$, $D_e$, $d_e$, $a$, and $b$) against DFT-computed energies for a number of atom-atom pairs. Specifically, we compute these DFT energies for each atom-atom pair by performing electronic structure optimization using VASP 5.4.4 \cite{kresse1996efficient}. The calculations use the generalized gradient approximation (GGA) of Perdew, Burke, and Ernzerhof (PBE) \cite{perdew_generalized_1996}. We apply an energy cutoff for the set of plane wave basis sets of 520 eV, and the energy tolerance is $10^{-6}$ eV. For each atom-atom pair, we perform the DFT optimization at 20 separation distances, and then fit Eq. \ref{eq:morse} against the energies obtained at these separation distances.

\textbf{Physics-Informed Loss Function}
% \textcolor{blue}{Hoang should revise this}
For a given molecular system at step $t$, the model predicts atomic displacements $\Delta_{t}$, which are used to update the atomic coordinates:
\begin{equation}
\mathbf{r}_{t+1} = \mathbf{r}_t + \Delta_{t}.
\end{equation}

We then sample \( M \) pairs of atoms \( \{(i_m, j_m)\}_{m=1}^M \) and compute their pairwise distances using the predicted positions:
\begin{equation}
\label{distance_eq}
d_{t+1}^{i_m,j_m} = \left\| \mathbf{r}_{t+1}^{i_m} - \mathbf{r}_{t+1}^{j_m} \right\|_2.
\end{equation}
To encourage physically plausible predictions at step \(t\), we define a physics‐informed loss based on the Morse potential:
\begin{equation}
\label{physic_loss}
\mathcal{L}_{\mathrm{phys}}
\;=\; 
\frac{1}{\lvert \mathbf{M}'\rvert}
\sum_{(i_m,j_m)\in M'} 
E\bigl(d_{t+1}^{\,i_m,j_m}\bigr),
\end{equation}
where
\begin{equation}
\mathbf{M}' 
= 
\bigl\{\, (i_m,j_m)\in \{(i_m,j_m)\}_{m=1}^M 
\;\bigm|\; 
E\bigl(d_{t+1}^{\,i_m,j_m}\bigr)
>
\tau_{\mathrm{train}}^{\,i,j} 
\bigr\}.
\end{equation}
Here, \(E(d)\) is the Morse potential energy calculated using Eq.~\ref{eq:morse} and \(\tau_{\mathrm{train}}^{\,i,j}\) is a predefined energy threshold for a pair of atoms \((i,j)\). The core idea is to selectively apply the physics constraint only to atomic pairs whose predicted interactions violate physical plausibility, as determined by the energy threshold. We compute \(\tau_{\text{train}}^{i,j}\) from the training dataset. For each atomic pair \((i, j)\), we iterate over all time steps in the training trajectories and compute the Morse potential energy \(E(d_t^{ij})\) at each step. The threshold \(\tau_{\text{train}}^{i,j}\) is then set as the maximum energy observed for that pair across the entire training set:
\begin{equation}
\label{tau_training}
\tau_{\text{train}}^{i,j} = \max_{t \in \text{Train}} E(d_t^{i,j}).
\end{equation}

\textbf{Optimization and Training Details} The training objective consists of a weighted sum of the mean squared error (MSE) between predicted and ground-truth displacements, along with a physics-informed regularization term:

\begin{equation}
\mathcal{L} = \mathcal{L}_{\text{MSE}} + \lambda \mathcal{L}_{\text{phys}},
\end{equation}
where  $\lambda$ is the physics-informed hyperparameter controlling the trade-off between predictive accuracy and physical consistency. We optimize this objective using the Adam optimizer with a learning rate $\eta$ and a mini-batch size of $B$. The training process updates the model parameters $\theta$ via gradient descent:

\begin{equation}
\theta \leftarrow \theta - \eta \frac{\partial \mathcal{L}}{\partial \theta}.
\end{equation}
To ensure stability during training, we apply gradient clipping and use an adaptive learning rate schedule. The model is trained with early stopping based on validation loss. This loss-based approach encourages the model to learn to predict stable molecular trajectories.

\subsection{Physics-Informed Inference (PII)} \label{sec:pii}
Although the model is trained with a physics-informed loss that penalizes high-energy atomic states, this alone does not guarantee physically plausible predictions at inference time. The learned model may still occasionally generate displacements that result in unphysical atom-atom interactions, such as atoms collapsing too closely, especially when extrapolating to long trajectories or out-of-distribution inputs.

To address this, we introduce a physics-informed inference correction step as a safeguard during trajectory rollout. Specifically, after predicting the atomic displacements and updating positions, we check whether any interatomic distances violate physical constraints by computing their Morse potential energy. If the predicted configuration results in excessive potential energy for any sampled pair, we discard the displacement and freeze the atoms' positions at the current step.

In particular, at each autoregressive step, similar to constructing the physics-informed loss, we  sample a set of atomic pairs $(i, j)$ and compute the corresponding interatomic distances $d_{t+1}^{i,j}$ using the updated coordinates:
\begin{equation}
d_{t+1}^{i,j} = || \mathbf{r}_{t+1}^i - \mathbf{r}_{t+1}^j ||_2.
\end{equation}
The Morse potential energy for each pair is evaluated as
\begin{equation}
E(d_{t+1}^{i,j}) = D_e \left(1 - e^{-a(d_{t+1}^{i,j} - d_e)}\right)^2 + b.
\end{equation}

To avoid the generation of unphysical configurations, we again apply the energy threshold \( \tau_{train}^{i,j} \) in Eq.~\ref{tau_training}. If the computed energy \( E(d_{t+1}^{i,j}) \) for any sampled pair exceeds its corresponding threshold \( \tau_{train}^{i,j} \), the predicted displacement \( \Delta_{t+1} \) is rejected and replaced with a zero vector:

\begin{equation}
\label{train_tau}
\Delta_{t+1} = 
\begin{cases}
\Delta_{t+1}, & \text{if } E(d_{t+1}^{i,j}) < \tau_{train}^{i,j} \text{ for all sampled pairs} \\
\mathbf{0}, & \text{otherwise}
\end{cases}
\end{equation}

This correction serves as a real-time filter that enforces conservative updates, effectively preventing the propagation of physically invalid configurations. This is particularly important during autoregressive inference at test time, where error accumulation can lead to out-of-distribution inputs, making the training-time loss in Eq. \ref{physic_loss} insufficient to guarantee constraint satisfaction. Inference pseudocode is provided in Algo. ~\ref{algo:pii}, with a runtime complexity that scales linearly with the prediction length $T$.

% \begin{algorithm}[H]
% \caption{Physics-Informed Autoregressive Inference}
% \begin{algorithmic}[1]
% \Require Initial atomic positions $\mathbf{r}_0 = \{ \mathbf{r}_0^i \}_{i=1}^N$, $\Delta \mathbf{r}_{-1}=\mathbf{0}$, forecasting model $f_\theta$, energy threshold $\tau$, number of steps $T$
% \Ensure Forecasted trajectory $\{\mathbf{r}_t\}_{t=1}^{T}$

% \For{$t = 0$ to $T-1$}
%     \State Construct $X_t = [\mathbf{r}_t, \Delta \mathbf{r}_{t-1}]$
%     \State Predict displacements: $\Delta \mathbf{r}_{t+1} = f_\theta(\mathbf{r}_t)$
%     \State Update positions: $\mathbf{r}_{t+1} = \mathbf{r}_t + \Delta \mathbf{r}_{t}$
%     \State Sample atomic pairs $\mathcal{P} = \{(i,j)\}$ from $1 \leq i,j \leq N$
%     \State Set $\texttt{violation} \gets \texttt{False}$
%     \For{each $(i,j) \in \mathcal{P}$}
%         \State Compute distance: $d_{t+1}^{ij} = \| \mathbf{r}_{t+1}^i - \mathbf{r}_{t+1}^j \|_2$
%         \State Compute energy: $E(d_{t+1}^{ij})$ as in Eq. \ref{eq:morse}. 
%         \If{$E(d_{t+1}^{ij}) > \tau$}
%             \State $\texttt{violation} \gets \texttt{True}$
%             \State \textbf{break}
%         \EndIf
%     \EndFor
%     \If{$\texttt{violation}$}
%         \State $\Delta \mathbf{r}_{t+1} \gets \mathbf{0}$
%         \State $\mathbf{r}_{t+1} \gets \mathbf{r}_t$
%     \EndIf
% \EndFor
% \end{algorithmic}
% \end{algorithm}

\begin{algorithm}[t]\label{algo:pii}
\caption{Physics-Informed Inference}
\begin{algorithmic}[1]
\Require Initial atomic positions $\mathbf{r}_1$, $\Delta_{0} = \mathbf{0}$, forecasting model $f_\theta$, energy threshold $\tau_{train}^{i,j}$, total steps $T$, forecasting window size $L$
\Ensure Forecasted trajectory $\{ \mathbf{r}_t \}_{t=1}^T$

\For{$t = 1$ to $T$}
    \State Construct input $X_t = [\mathbf{r}_t, \Delta_{t-1}]$
    \State Predict future displacements:  $\Delta_{t:t+L-1} = f_\theta(X_t)$
    \For{$\ell = 0$ to $L-1$}
        \State Compute $\mathbf{r}_{t+\ell+1} = \mathbf{r}_{t+\ell} + \Delta_{t+\ell}$
        \State Sample atomic pairs $\mathcal{P} = \{(i,j)\}$ from $[1, N]$
        \State $\texttt{violation} \gets \texttt{False}$
        \For{each $(i,j) \in \mathcal{P}$}
            \State Compute $d_{t+\ell}^{i,j} = \| \mathbf{r}_{t+\ell}^i - \mathbf{r}_{t+\ell}^j \|_2$
            \State Compute $E(d_{t+\ell}^{i,j})$ using Morse potential
            \If{$E(d_{t+\ell}^{i,j}) > \tau_{train}^{i,j}$}
                \State $\texttt{violation} \gets \texttt{True}$
                \State \textbf{break}
            \EndIf
        \EndFor
        \If{$\texttt{violation}$}
            \State $\Delta_{t+\ell} \gets \mathbf{0}$
            \State $\mathbf{r}_{t+\ell+1} = \mathbf{r}_{t+\ell} + \Delta_{t+\ell}$
        \EndIf
    \EndFor
\EndFor
\end{algorithmic}
\end{algorithm}

%% file: p_exp.tex
\subsection{Generation of AIMD Atomic Data} 
To create training and testing data for machine learning models, we carried out AIMD simulations using the Vienna Ab initio Simulation Package (VASP)~\cite{kresse1996efficient}. We simulated a diverse set of materials containing different atomic species, with system sizes up to 300 atoms. These simulations were run at temperatures ranging from 600\,K to 1200\,K. At these temperatures, atomic vibrations become significant, making it possible to probe various thermodynamic behaviors. However, the resulting complex and noisy trajectories present a major challenge for learning-based models aiming to accurately predict molecular dynamics over time.

Each simulation used a time step of 1~fs, updating atomic positions and velocities via the Verlet integration scheme~\cite{grubmuller1991generalized}. All simulations were performed in the NVT ensemble using the Nosé thermostat to maintain constant temperature. The atomic interactions were modeled using a projector-augmented wave (PAW) pseudopotential, along with the Perdew–Burke–Ernzerhof (PBE) exchange-correlation functional under the generalized gradient approximation (GGA). A plane-wave energy cutoff of 500~eV and single $\Gamma$-point sampling was used.

The simulation produces six datasets, each corresponding to a distinct combination of temperature and sequence length. For each dataset, we apply a train/validation/test split and summarize the resulting dataset statistics in Table \ref{tab:dataset_stats}.

\begin{table}[t]
\centering
\caption{Dataset statistics across different temperatures and materials.}
\begin{tabular}{|l|c|c|c|}
\hline
\textbf{Dataset} & \textbf{Material} & \textbf{Temperature (K)} & \textbf{Train/Valid/Test Size} \\
\hline
A & Li$_{10}$GeP$_2$S$_{12}$ & 600  & 28,000 / 4,000 / 1,000 \\
\hline
B & Li$_{10}$GeP$_2$S$_{12}$ & 800  & 28,000 / 4,000 / 1,000 \\
\hline
C & Li$_{10}$GeP$_2$S$_{12}$ & 1,000 & 28,000 / 4,000 / 1,000 \\
\hline
D & Li$_{10}$GeP$_2$S$_{12}$ & 1,200 &  28,000 / 4,000 / 1,000 \\
\hline
E & Li\textsubscript{7}Ta\textsubscript{1}O\textsubscript{6} & 800  &  28,000 / 4,000 / 1,000 \\
\hline
F & Li\textsubscript{7}Ta\textsubscript{1}O\textsubscript{6} & 1,000 & 19,144 / 2,735 / 1,000 \\
\hline
G & LiGaBr\textsubscript{3} & 1,000 & 5,967 / 852 / 1,000 \\
\hline
\end{tabular}
\label{tab:dataset_stats}
\end{table}

\subsection{Training and Evaluation Setup}
All models were trained using a single NVIDIA A100 GPU, with training times ranging from 1 to 1.6 hours depending on the dataset size. To assess result variability, we initially conducted multiple training runs and measured the mean and standard deviation of the final performance metrics. However, we observed that the variations were negligible. Therefore, in line with common practice in time-series forecasting, we report results from a single representative training run.

To evaluate each model, we begin with an initial window of atomic positions of length $H$. Using this as input, the model autoregressively predicts the future atomic trajectory over a specified forecast horizon (1000 testing steps). Importantly, during this prediction phase, the model does not receive ground-truth positions at each step. Instead, it feeds its own previously predicted positions back as input for subsequent predictions. This setup reflects a more realistic and challenging scenario, as it simulates deployment conditions where ground-truth data is unavailable beyond the observed history.

\textbf{Forecasting metrics} To assess how close the predicted atomic trajectories are to the ground-truth data, we adopt standard metrics from the time-series forecasting literature, namely the Mean Squared Error (MSE) and Mean Absolute Error (MAE). These metrics can be computed between the predicted and ground-truth values, which can refer to either atomic displacements \(\Delta_t^i \in \mathbb{R}^3\) or positions \(\mathbf{r}_t^i \in \mathbb{R}^3\). For a forecast horizon of length \(L\) over \(N\) atoms, the metrics are:

\paragraph{Displacement}
\begin{equation}
\text{MSE}_{\Delta} = \frac{1}{L N} \sum_{t=1}^{L} \sum_{i=1}^{N} \left\| \hat{\Delta}_t^i - \Delta_t^i \right\|_2^2,
\end{equation}
\begin{equation}
\text{MAE}_{\Delta} = \frac{1}{L N} \sum_{t=1}^{L} \sum_{i=1}^{N} \left\| \hat{\Delta}_t^i - \Delta_t^i \right\|_2,
\end{equation}

\paragraph{Position}
\begin{equation}
\text{MSE}_{\mathbf{r}} = \frac{1}{L N} \sum_{t=1}^{L} \sum_{i=1}^{N} \left\| \hat{\mathbf{r}}_t^i - \mathbf{r}_t^i \right\|_2^2,
\end{equation}
\begin{equation}
\text{MAE}_{\mathbf{r}} = \frac{1}{L N} \sum_{t=1}^{L} \sum_{i=1}^{N} \left\| \hat{\mathbf{r}}_t^i - \mathbf{r}_t^i \right\|_2.
\end{equation}

These metrics quantify the average squared and absolute deviations, serving as reliable indicators of forecasting accuracy for both displacement and position predictions.

\textbf{Physical metrics} To evaluate the physical plausibility of predicted trajectories, we introduce physical metrics that quantify how often the model produces energetically infeasible atomic configurations. Specifically, for each forecasted step $t$, we sample a subset of atomic pairs $(i,j)$ and compute their pairwise distances \(d_t^{ij} = \| \mathbf{r}_t^i - \mathbf{r}_t^j \|_2\), which are then used to calculate the corresponding Morse potential energies \(E(d_t^{ij})\).

A violation is recorded if the energy between any sampled pair exceeds a threshold \(\tau_{i,j}\). We define a binary indicator:
\begin{equation}
\label{test_tau}
\mathbb{I}_t^{ij} = 
\begin{cases}
1, & \text{if } E(d_t^{ij}) > \tau_{i,j} \\
0, & \text{otherwise}
\end{cases}
\end{equation}
Here, \(\tau_{i,j}\) can be estimated using the test dataset as follows: for each atomic pair \((i,j)\), we iterate over all time steps in the test trajectories and record the Morse potential energy \(E(d_t^{ij})\) at each step. The threshold \(\tau_{i,j}\) is then set to the maximum energy observed for that pair across the entire test set:
\begin{equation}
\tau_{i,j} = \max_{t \in \text{Test}} E(d_t^{ij}).
\end{equation}

This data-driven approach ensures that the violation criterion reflects the highest physically valid interaction energy encountered under ground-truth dynamics. By using pair-specific thresholds instead of a global constant, we accommodate natural variations in interaction strength and equilibrium distances across different atom types or bonded states.

Let \(\mathcal{P}_t\) denote the set of sampled atomic pairs at time \(t\) with $|\mathcal{P}_t|=M$, the \textit{total number of violations} $V_{n}$ over a forecast horizon of length \(L\) is computed as:
\begin{equation}
V_n(L,M) = \sum_{t=1}^{L} \sum_{(i,j) \in \mathcal{P}_t} \mathbb{I}_t^{ij}.
\end{equation}

To allow for a fair comparison across different systems and trajectories, we also report the normalized \textit{violation rate} $V_r$, which captures the average proportion of violating pairs per time step:
\begin{equation}
V_r = \frac{1}{LM} V_n.
\end{equation}

These violation-based metrics are orthogonal to traditional error metrics and serve as a proxy for the degree of physical constraint adherence in the predicted dynamics.

\begin{table}[t]
\centering
\caption{Performance comparison between our method \textit{PhyTimeMD} and conventional MD baselines. Lower values of $\text{MSE}_\mathrm{r}$, $\text{MAE}_\mathrm{r}$, and \(V_r\) indicate better performance. Best results are \textbf{bolded}.
}
\begin{tabular}{|l|l|c|c|c|c|}
\hline
\textbf{Dataset} & \textbf{Model}   & \textbf{$\text{MSE}_\text{r}$} & \textbf{$\text{MAE}_\text{r}$} & \textbf{$V_r \%$} \\
\hline
 & LSTM &  $30.57$ & $4.65$ & $33.833$\\
 A & TimeMixer & $10.08$  & $2.05$& $3.384$ \\
 & PhysTimeMD (Ours)  & $\textbf{6.20}$ & $\textbf{0.98}$ & $\textbf{0.028}$ \\
 \hline
 & LSTM  & $27.14$ & $4.36$ & $31.074$\\
 B & TimeMixer  & $13.67$ & $2.66$ & $2.677$ \\
 & PhysTimeMD (Ours) & $\textbf{6.78}$ & $\textbf{1.13}$ & $\textbf{0.034}$ \\
 \hline
  & LSTM  & $31.65$ & $4.71$ & $33.754$\\
 C & TimeMixer  &  $12.52$ & $2.53$ & $5.410$ \\
 & PhysTimeMD (Ours)  & $\textbf{5.92}$ & $\textbf{1.11}$ & $\textbf{0.000}$ \\
 \hline
  & LSTM  &  $31.17$ & $4.64$ & $27.797$\\
 D & TimeMixer   & $17.78$ & $3.89$ & $4.110$ \\
 & PhysTimeMD (Ours)  & $\textbf{6.08}$ & $\textbf{1.13}$ & $\textbf{0.026}$ \\
\hline
 & LSTM   & $32.35$ & $4.40$ & $44.090$\\
 E& TimeMixer&  $10.35$ & $1.78$ & $9.984$ \\
 & PhysTimeMD (Ours)  & $\textbf{4.49}$ & $\textbf{0.82}$ & $\textbf{6.760}$ \\
\hline
 & LSTM  &  $33.39$ & $4.48$ & $42.164$\\
 F & TimeMixer   & $7.63$ & $1.25$ & $6.725$\\
 & PhysTimeMD (Ours)  & $\textbf{3.76}$ & $\textbf{0.77}$ & $\textbf{2.310}$ \\
\hline
& LSTM   & $52.92$ & $6.09$ &$31.237$\\
 G& TimeMixer  &  $31.38$ & $4.51$ & $7.578$\\
 & PhysTimeMD (Ours)  & $\textbf{9.54}$ & $\textbf{1.18}$ & $\textbf{1.140}$ \\
\hline
\end{tabular}

\label{tab:conventional_baselines}
\end{table}

\subsection{Comparison with Conventional MD models}
To evaluate the effectiveness of our approach, we compare it against conventional MD models that typically predict future atomic positions directly from past positions without any physics constraint. Specifically, we follow prior work on modeling MD trajectories~\cite{eslamibidgoli2019recurrent, tsai2020learning, wang2020accelerated} and implement an LSTM network for position prediction. We also include an additional baseline based on TimeMixer~\cite{wang2024timemixer}, a strong deep-learning model for time-series forecasting. This baseline represents a straightforward approach that directly applies a time-series model to predict atomic positions. For a fair comparison, our PhysTimeMD framework also adopts TimeMixer as the forecast backbone. We evaluate the models using $\text{MSE}_\text{r}$ and $\text{MAE}_\text{r}$. It is worth noting that generative and graph methods \cite{endo2018multi, nam2024flow, sibi2024advancing} cannot apply to our datasets, as they rely on prior structural information that is unavailable in our setting.

The results in Table~\ref{tab:conventional_baselines} show that our PhysTimeMD framework consistently outperforms both LSTM and TimeMixer across all datasets. For example, \textbf{in dataset C}, PhysTimeMD achieves an $\text{MSE}_\text{r}$ of $5.92$ and an $\text{MAE}_\text{r}$ of $1.11$, compared to $12.52$ and $2.53$ for TimeMixer, and $31.65$ and $4.71$ for LSTM, respectively. The violation rate $V_r$ is reduced to $0.0\%$, in stark contrast to $5.4\%$ for TimeMixer and $33.7\%$ for LSTM. Similar improvements are observed across the other datasets. \textbf{In dataset E}, PhysTimeMD achieves an $\text{MSE}_\text{r}$ of $4.49$ and $\text{MAE}_\text{r}$ of $0.82$, improving upon TimeMixer ($\text{MSE}_\text{r}$: $10.35$, $\text{MAE}_\text{r}$: $1.78$) and LSTM ($\text{MSE}_\text{r}$: $32.35$, $\text{MAE}_\text{r}$: $4.40$). \textbf{In dataset G}, where the baseline errors are particularly high, PhysTimeMD reduces the $\text{MSE}_\text{r}$ from $52.92$ (LSTM) and $31.38$ (TimeMixer) to just $9.54$, and $\text{MAE}_\text{r}$ from $6.09$ and $4.51$ down to $1.18$.

Moreover, our violation rate $V_r$ is consistently reduced by at least an order of magnitude. LSTM models yield $V_r$ values between $27.8\%$ and $44.1\%$, while TimeMixer achieves $2.7\%$ to $9.9\%$. In contrast, PhysTimeMD keeps $V_r$ below $1.2\%$ across all datasets. For example, $V_r$ drops to $0.028\%$ \textbf{in dataset A}, $0.034\%$ in B, and $0.026\%$ in D. These results confirm the effectiveness of our approach, which combines displacement modeling with physics-informed components.

\begin{figure*}[t]
    \centering
    \includegraphics[width=0.92\linewidth]{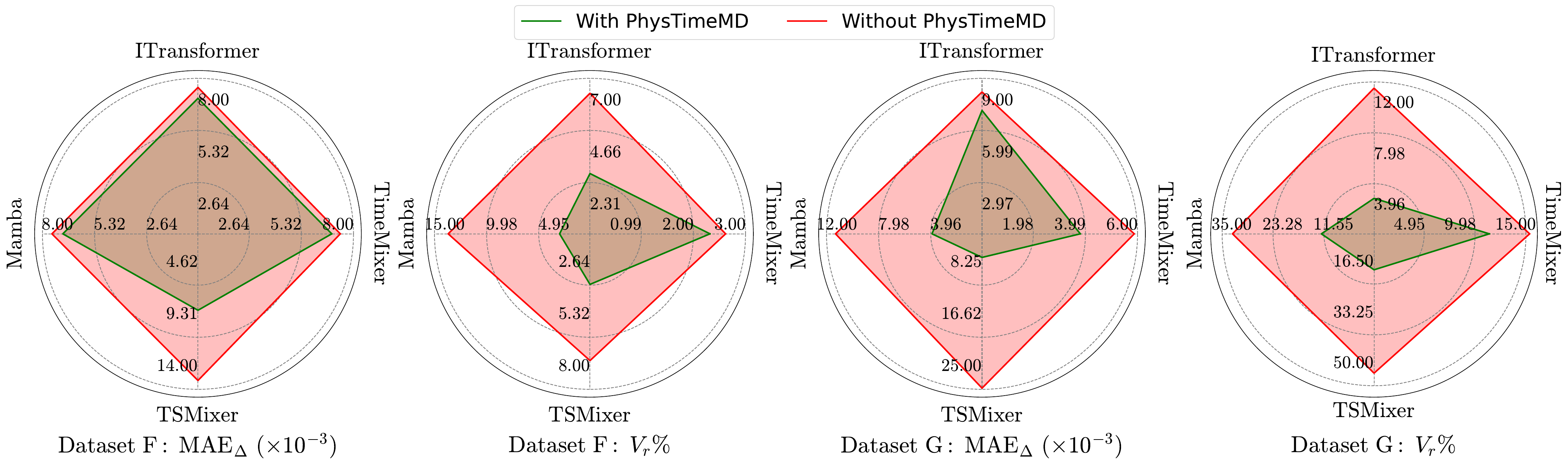}
    \caption{Radar plot comparing forecasting error ($\text{MAE}_\Delta$) and normalized physical violation rate \(V_r\) (\%) across datasets G (left) and F (right). The lower the metrics the better the results. Each axis represents one of the four baseline architectures—TimeMixer, ITransformer, Mamba, and TSMixer—with and without the PhysTimeMD framework, highlighting improvements in both accuracy and physical consistency.}
    \label{fig:timeseries_baselines}
\end{figure*}

\subsection{Benchmarking with Time-Series Baselines}
In this section, we examine our PhysTimeMD as a general framework that can be applied to any modern time-series model to equip them with physics-informed components. To create an equal impact of displacement modeling, we apply the Displacement-based Time-Series Forecasting Formulation (Sec. \ref{sec:framework}) uniformly across all forecasting models. Under this setup, we evaluate several state-of-the-art time-series architectures both with and without the integration of PhysTimeMD. These include \textit{TimeMixer}~\cite{wang2024timemixer}, \textit{ITransformer}~\cite{liu2024itransformer}, \textit{Mamba}~\cite{gu2024mamba}, and \textit{TSMixer}~\cite{chen2024tsmixer}. We implement these baselines using a public code repository introduced in \cite{wang2024tssurvey}.

For illustrative purposes, Fig.~\ref{fig:timeseries_baselines} reports $\text{MAE}_\Delta$ and normalized physical violation rate \(V_r\) (in \%) for datasets G and F, while the comprehensive results are provided in Appendix
Tab.~\ref{tab:timeseries_baselines} with all datasets and metrics. \textbf{In dataset F}, all architectures benefit from PhysTimeMD: the $\text{MAE}_\Delta$ of TSMixer decreases from \(1.32\times10^{-2}\) to \(6.88\times10^{-3}\) with \(V_r\) halved (from \(6.5\%\) to \(2.6\%\)), and ITransformer, despite only a modest improvement in $\text{MAE}_\Delta$ (from \(7.53\times10^{-3}\) to \(6.97\times10^{-3}\)), exhibits a substantial decrease in violation rate (from \(6.3\%\) to \(2.7\%\)). \textbf{In dataset G}, PhysTimeMD yields remarkable improvements: the $\text{MAE}_\Delta$ of TimeMixer decreases from \(5.87\times10^{-3}\) to \(3.79\times10^{-3}\) with \(V_r\) reduced from \(15.4\%\) to \(11.4\%\); TSMixer’s $\text{MAE}_\Delta$ is reduced fivefold (from \(2.48\times10^{-2}\) to \(3.79\times10^{-3}\)) while \(V_r\) declines nearly fourfold (from \(45.8\%\) to \(11.8\%\)). Two notable cases further highlight the benefits of PhysTimeMD. \textbf{In dataset B}, although TimeMixer without PhysTimeMD slightly outperforms the version with PhysTimeMD in terms of $\text{MAE}_\Delta$ and $\text{MSE}_\Delta$, its violation rate is \(6.1\%\) compared to \(0.034\%\) of PhysTimeMD, rendering it impractical to be used in reality. This underscores the capability of PhysTimeMD to enforce physical consistency independently of the forecast error. \textbf{In dataset E}, the backbone TSMixer without PhysTimeMD diverges, producing unbounded position predictions that preclude reliable MAE/MSE computation, whereas with PhysTimeMD it remains stable (see Appendix~\ref{appendix:divergence_analysis} for an analysis of this). Across all remaining datasets, PhysTimeMD consistently enhances both forecasting accuracy and adherence to physical constraints, demonstrating that integrating physics‐informed methodologies into contemporary time series models yields predictions that are both more precise and inherently physically plausible.

\subsection{Ablation Study and Hyperparameter Analysis}
\textbf{Effectiveness of Physics-Informed Training and Inference.}
In this experiment, we investigate the contributions of the individual components of PhysTimeMD, specifically Physics-Informed Training (PIT) and Physics-Informed Inference (PIF). The evaluation is conducted using the TimeMixer backbone on Dataset A. The hyperparameter $\lambda$ is fixed at 0.0001, as this value yields the best performance (refer to Tab.~\ref{tab:timeseries_baselines}). The ablation results in Tab.~\ref{tab:ablation_physmd} demonstrate that both PIT and PIF significantly enhance the TimeMixer backbone’s performance. Enabling PIT alone reduces MAE from \(6.05\times10^{-3}\) to \(5.73\times10^{-3}\) (a \(5.3\%\) improvement), MSE (a \(9.0\%\) improvement), and \(V_r\) (a \(35.1\%\) reduction). Enabling PIF alone yields larger gains: MAE decreases with an improvement of \(21.2\%\), MSE (\(36.6\%\) improvement), and \(V_r\) (a \(99.0\%\) reduction). Combining both PIT and PIF (full PhysTimeMD) achieves the best overall results, with MAE \(4.60\times10^{-3}\) (a \(24.0\%\) improvement), MSE \(3.82\times10^{-5}\) (a \(40.9\%\) improvement), and \(V_r\) \(0.028\%\) (a \(99.4\%\) reduction). These findings confirm that each component contributes uniquely to accuracy and physical consistency, and their joint application delivers the best results.
\begin{table}[t]
\centering
\caption{Ablation study of PhysTimeMD components—Physics-Informed Training (PIT) and Physics-Informed Inference (PIF)—using the TimeMixer backbone on Dataset A. Reported metrics include $\text{MAE}_{\Delta}$, $\text{MSE}_{\Delta}$, and the physical violation rate $V_r$ (\%).}
\begin{tabular}{@{}ccccc@{}}
\hline
\textbf{PIT} & \textbf{PIF} & $\text{MAE}_{\Delta}$ & $\text{MSE}_{\Delta}$ & \textbf{$V_r$ \%} \\ 
\hline
\xmark & \xmark & $6.05\times10^{-3}$ & $6.47\times10^{-5}$ & $4.36$ \\
\cmark & \xmark & $5.73\times10^{-3}$ & $5.89\times10^{-5}$ & $2.83$ \\
\xmark & \cmark & $4.77\times10^{-3}$ & $4.10\times10^{-5}$ & $0.045$ \\
\cmark & \cmark & \bm{$4.60\times10^{-3}$} & \bm{$3.82\times10^{-5}$} & \bm{$0.028$} \\
\hline
\end{tabular}
\label{tab:ablation_physmd}
\end{table}

\begin{figure}[t]
    \centering
    \includegraphics[width=0.7\linewidth]{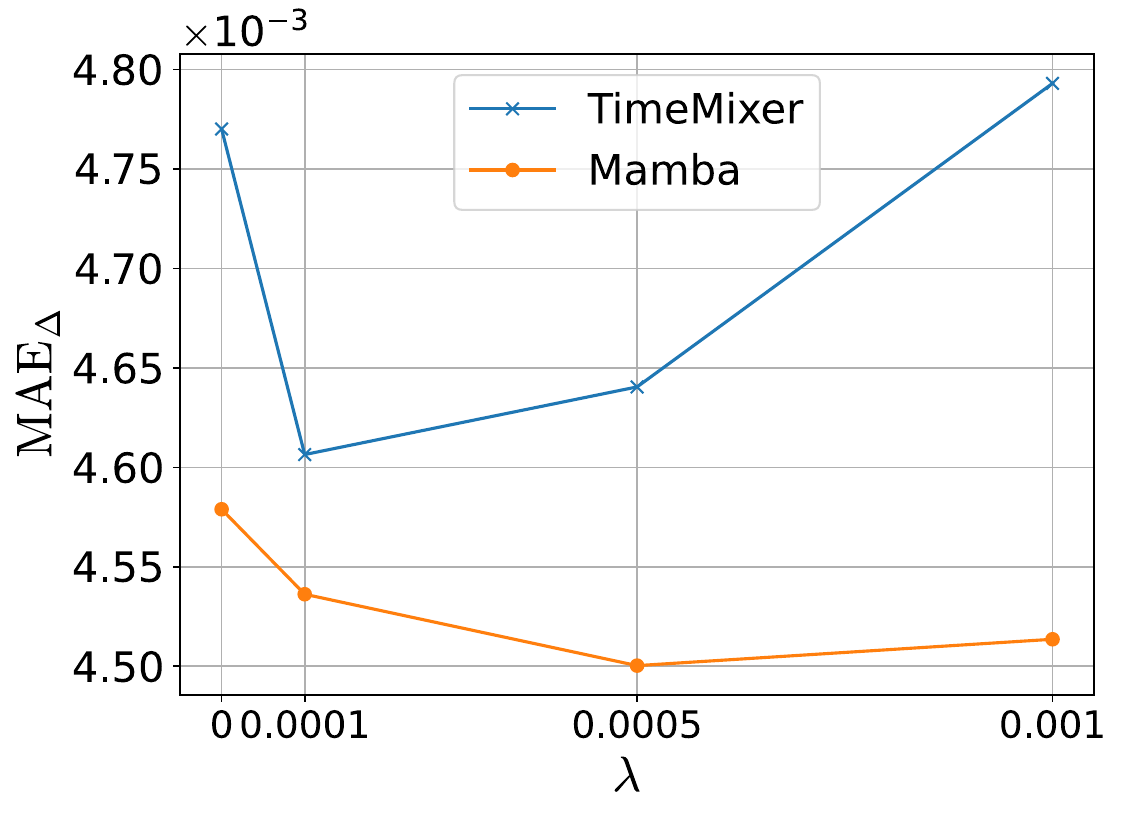}
    \caption{Effect of \(\lambda\) on $\text{MAE}_\Delta$ performance for TimeMixer and Mamba on Dataset A. Each curve illustrates how careful tuning of this hyperparameter can positively influence prediction accuracy.}
    \label{fig:lambda_2_tuning}
\end{figure}

\textbf{Sensitivity to the Hyperparameter \(\lambda\)}
We perform an ablation study to examine the effect of the physics-based hyperparameter \(\lambda\) on model accuracy, as shown in Fig.~\ref{fig:lambda_2_tuning}. We explore three tuned values—\(0.0001\), \(0.0005\), and \(0.001\)—alongside a value of \(\lambda = 0\), which corresponds to removing the physics-informed training component. Both models show improved performance within the tuned range. Specifically, Mamba achieves its lowest MAE of $4.51 \times 10^{-3}$ at \(\lambda = 0.0005\), while TimeMixer performs best at \(\lambda = 0.0001\) with a MAE of $4.60 \times 10^{-3}$. These findings indicate that the hyperparameter \(\lambda\) plays a critical role in model accuracy; however, excessively large values may degrade performance.

\subsection{Evaluation of Material Properties}

To assess the physical consistency of the predicted atomic trajectories, we compute the \textbf{diffusion coefficient} \( D \), which captures how atoms diffuse over time. Diffusivity is a key material property related to ionic transport, defect motion, and phase behavior in solids and fluids. We estimate \( D \) using the Einstein relation:

\begin{equation}
\left\langle \left| \mathbf{r}(t) - \mathbf{r}(0) \right|^2 \right\rangle = 2nDt,
\end{equation}
where \( \mathbf{r}(t) \) is the position of an atom at time \( t \), \( n \) is the number of spatial dimensions, and \( \left\langle \cdot \right\rangle \) denotes an average over atoms and simulation time. While it is important for a model to accurately predict the positions of atoms in a dynamics simulation, it is more important for the model to satisfy key quantitative and qualitative features, such as the diffusivity of ions in the material.

\begin{figure*}[t]
    \centering
    \includegraphics[width=1\linewidth]{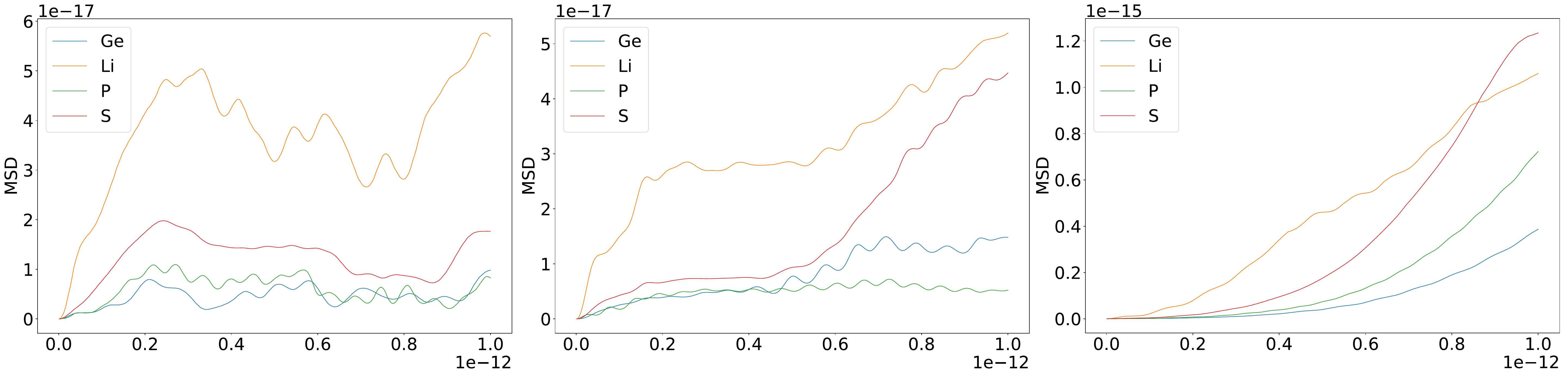}
    \caption{Diffusivity computed from ground-truth trajectories (left), PhysTimeMD-generated trajectories (middle), and TimeMixer-generated trajectories (right).
}
    \label{fig:diff}
\end{figure*}

In Fig. \ref{fig:diff}, we report the diffusivity estimated from our model's generated trajectories and compare it against that from a strong time-series forecaster (TimeMixer) and ground-truth data of dataset A. Closer agreement with the ground-truth indicates stronger physical fidelity in the long-term behavior of the predictions. PhysTimeMD achieves a diffusivity magnitude around $10^{-17}$, closely matching the ground-truth range, 
while the baseline model TimeMixer significantly overestimates diffusivity at approximately $10^{-15}$. 
More importantly, all the baseline curves exhibit a consistent upward drift over time, reflecting crystal melting. 
In contrast, PhysTimeMD maintains stable diffusion trends, with only two cases showing minor deviation (Ge and S), compared to just one 
in the ground truth (Ge). This demonstrates that PhysTimeMD not only captures accurate and stable long-term atomic dynamics but also generates trajectories with realistic diffusivity values, making them a viable replacement for ground-truth data in downstream material simulations.

Regarding running time, our method can generate 1,000 simulation steps in approximately 6 minutes on a single NVIDIA A100 GPU. In contrast, DFT-based molecular dynamics simulations are significantly more computationally intensive, requiring approximately 24 hours on a 96-CPU core computational node. This highlights the substantial efficiency advantage of our approach over traditional DFT simulations. To reproduce the results, we will release the source code and data upon publication.

%% file: p_related.tex
Machine learning has emerged as a powerful paradigm for enhancing MD simulations, addressing the long-standing trade-off between simulation accuracy and computational cost. Early machine learning approaches, such as support vector machines and autoencoders, were primarily used to extract thermodynamic properties like free energy from MD trajectories and to approximate full Boltzmann distributions for small biomolecules \cite{noe2019boltzmann}. However, these methods fall short of capturing high-resolution femtosecond (fs) dynamics—one of the central goals of MD simulations.

Recent advances in machine learning have significantly enhanced the modeling and acceleration of MD simulations. Broadly, three classes of deep learning approaches have emerged as particularly influential: recurrent neural networks, generative models, and graph neural networks.

Recurrent neural networks (RNNs), including LSTMs, have been applied to model MD trajectories as high-frequency, high-dimensional time series data \cite{eslamibidgoli2019recurrent, tsai2020learning, wang2020accelerated}. These models can capture short-term dynamics but often struggle with long-range dependencies and scalability, especially when modeling complex biomolecular systems with more than a few dozen particles. This limitation stems from their reliance on predicting atomic coordinates directly, without integrating physical principles, which leads to error accumulation and poor generalization.

Generative models—such as GANs, diffusion, and flow-based approaches—have been explored to produce long MD trajectories by stitching together short segments \cite{endo2018multi, schreiner2023implicit, jing2024generative, fu2024latent, zheng2024predicting, nam2024flow}. While these methods show promise, they often suffer from exposure bias, where small errors compound over iterations, leading to unrealistic long-term dynamics. Training generative models is often unstable and data-hungry, requiring careful balancing of adversarial losses or diffusion schedules, and they lack built-in physical constraints, making it difficult to ensure energy conservation or adherence to molecular symmetries. This limits their reliability for simulating fine-grained, high-resolution trajectories over extended timescales.

Graph neural networks (GNNs) offer a compelling alternative by modeling molecular systems as graphs, with atoms as nodes and interactions as edges \cite{sibi2024advancing, zheng2021learning, maji2025accelerating}. GNNs excel at capturing both local atomic environments and global physical effects and eliminate the need for handcrafted features through automatic representation learning. However, GNNs also present challenges: they require careful graph construction based on spatial proximity, which may be unstable at high temperatures or during rapid conformational changes, and their complex architectures demand extensive tuning and computational resources. Given these limitations, modeling MD as a standard time-series forecasting provides a more tractable and scalable framework.

%% file: p_discuss.tex
In this paper, we introduce a displacement-based, physics-informed forecasting framework for molecular dynamics. By predicting atomic displacements and incorporating the Morse potential into both training and inference, our method achieves stable and physically plausible simulations over long horizons while remaining computationally efficient. Experimental results on various materials at different temperatures show that our approach not only improves the accuracy of state-of-the-art time-series forecasting models for molecular dynamics but also outperforms conventional MD methods in both precision and efficiency, producing stable trajectories for up to 1,000 simulation steps. Although this marks a significant improvement, small prediction errors still accumulate over time, gradually degrading physical fidelity. Future work will aim to address this by extending the stability and accuracy of the framework to enable robust long-range simulations over 10,000 steps.

%% file: p_appendix.tex
\begin{table*}[!tb]
\centering
\caption{Comparison of model performance with and without PhysTimeMD on multiple datasets. For each model, we select the $\lambda$ value that yields the lowest $\text{MAE}_{\Delta}$ and then report that MAE alongside corresponding, $\text{MAE}_\mathrm{r}$, $\text{MSE}_{\Delta}$, $\text{MSE}_\mathrm{r}$ and its corresponding $V_r$\%. In this experiment, we calculate the displacements and convert it back into the raw positions to calculate $\text{MAE}_\mathrm{r}$ and $\text{MSE}_\mathrm{r}$. An asterisk (*) denotes cases where the model diverged, making MAE, MSE, and $V_r\%$ unreliable. \textbf{Bold} values indicate improvements achieved by incorporating PhysTimeMD.}
\resizebox{\textwidth}{!}{
\begin{tabular}{|c|l|c|c|c|c|c|c|c|}
\hline
\textbf{Dataset} & \textbf{Model} & \textbf{PhysTimeMD} & $\boldsymbol{\lambda}$ & $\text{MAE}_{\Delta}$ & $\text{MAE}_\text{r}$ & $\text{MSE}_{\Delta}$ & $\text{MSE}_\text{r}$ & \textbf{$V_r$ \%} \\
\hline
\multirow{8}{*}{A}
 & TimeMixer     & \xmark & ---    & $6.05\times10^{-3}$ & 1.81 & $6.47\times10^{-5}$ & 9.62  & $4.36$  \\
 &               & \cmark & \textbf{0.0001} & \bm{$4.60\times10^{-3}$} & \textbf{0.98} & \bm{$3.82\times10^{-5}$} & \textbf{6.20}  & \bm{$0.028$}  \\
 & ITransformer  & \xmark & ---    & $5.94\times10^{-3}$ & 2.50 & $6.07\times10^{-5}$ & 13.23 & $6.08$  \\
 &               & \cmark & \textbf{0.0005} & \bm{$4.76\times10^{-3}$} & \textbf{1.01} & \bm{$4.00\times10^{-5}$} & \textbf{6.17}  & \bm{$0.005$}  \\
 & Mamba         & \xmark & ---    & $5.37\times10^{-3}$ & 2.04 & $5.25\times10^{-5}$ & 11.55 & $3.59$  \\
 &               & \cmark & \textbf{0.0005} & \bm{$4.51\times10^{-3}$} & \textbf{0.96} & \bm{$3.64\times10^{-5}$} & \textbf{6.11}  & \bm{$0.128$} \\
 & TSMixer       & \xmark & ---    & $6.91\times10^{-3}$ & 2.32 & $8.41\times10^{-5}$ & 13.78 & $8.62$  \\
 &               & \cmark & \textbf{0.001}  & \bm{$4.51\times10^{-3}$} & \textbf{1.10} & \bm{$3.68\times10^{-5}$} & \textbf{6.21}  & \bm{$0.002$}  \\
\hline
\multirow{8}{*}{B}
 & TimeMixer     & \xmark & ---    & $1.48\times10^{-3}$ & 0.89 & $6.76\times10^{-6}$ & 4.37 & $6.06$  \\
 &               & \cmark & 0.0005 & $5.24\times10^{-3}$ & 1.13 & $5.00\times10^{-5}$ & 6.78  & $0.034$ \\
 & ITransformer  & \xmark & ---    & $7.72\times10^{-3}$ & 2.55 & $1.05\times10^{-3}$ & 13.65 & $10.50$ \\
 &               & \cmark & \textbf{0.001}  & \bm{$5.53\times10^{-3}$} & \textbf{1.13} & \bm{$5.45\times10^{-5}$} & \textbf{6.73}  & \bm{$0.090$} \\
 & Mamba         & \xmark & ---    & $7.82\times10^{-3}$ & 2.58 & $1.09\times10^{-4}$ & 14.85 & $5.92$  \\
 &               & \cmark & \textbf{0.0005} & \bm{$6.47\times10^{-3}$} & \textbf{1.34} & \bm{$7.54\times10^{-5}$} & \textbf{6.61}  & \bm{$0.117$} \\
 & TSMixer       & \xmark & ---    & $7.75\times10^{-3}$ & 2.33 & $1.32\times10^{-4}$ & 15.20 & $11.20$ \\
 &               & \cmark & \textbf{0.001}  & \bm{$4.61\times10^{-3}$} & \textbf{1.12} & \bm{$3.87\times10^{-5}$} & \textbf{6.39}  & \bm{$0.058$} \\
\hline
\multirow{8}{*}{C}
 & TimeMixer     & \xmark & ---    & $7.52\times10^{-3}$ & 1.89 & $9.63\times10^{-5}$ & 9.12  & $1.95$  \\
 &               & \cmark & \textbf{0.0001} & \bm{$5.98\times10^{-3}$} & \textbf{1.11} & \bm{$6.55\times10^{-5}$} & \textbf{5.92}  & \bm{$0.000$} \\
 & ITransformer  & \xmark & ---    & $9.66\times10^{-3}$ & 3.34 & $1.50\times10^{-4}$ & 24.48 & $24.60$ \\
 &               & \cmark & \textbf{0.0005} & \bm{$6.06\times10^{-3}$} & \textbf{1.06} & \bm{$6.69\times10^{-5}$} & \textbf{5.94}  & \bm{$0.004$} \\
 & Mamba         & \xmark & ---    & $6.92\times10^{-3}$ & 2.14 & $8.44\times10^{-5}$ & 10.42 & $2.64$  \\
 &               & \cmark & \textbf{0.0001} & \bm{$5.89\times10^{-3}$} & \textbf{1.12} & \bm{$6.36\times10^{-5}$} & \textbf{5.69}  & \bm{$0.196$} \\
 & TSMixer       & \xmark & ---    & $1.33\times10^{-2}$ & 3.87 & $3.74\times10^{-4}$ & 40.46 & $17.70$ \\
 &               & \cmark & \textbf{0.0005} & \bm{$5.84\times10^{-3}$} & \textbf{1.19} & \bm{$6.31\times10^{-5}$} & \textbf{6.11}  & \bm{$0.073$} \\
\hline
\multirow{8}{*}{D}
 & TimeMixer     & \xmark & ---    & $7.47\times10^{-3}$ & 2.57 & $9.72\times10^{-5}$ & 14.69 & $8.08$  \\
 &               & \cmark & \textbf{0.001}  & \bm{$5.94\times10^{-3}$} & \textbf{1.13} & \bm{$6.51\times10^{-5}$} & \textbf{6.08}  & \bm{$0.026$} \\
 & ITransformer  & \xmark & ---    & $7.87\times10^{-3}$ & 2.43 & $1.06\times10^{-4}$ & 12.49 & $6.08$  \\
 &               & \cmark & \textbf{0.0001} & \bm{$6.60\times10^{-3}$} & \textbf{1.41} & \bm{$7.88\times10^{-5}$} & \textbf{6.99}  & \bm{$0.026$} \\
 & Mamba         & \xmark & ---    & $7.58\times10^{-3}$ & 2.54 & $9.99\times10^{-5}$ & 13.73 & $5.12$  \\
 &               & \cmark & \textbf{0.0005} & \bm{$6.48\times10^{-3}$} & \textbf{1.36} & \bm{$7.54\times10^{-5}$} & \textbf{6.73}  & \bm{$0.018$} \\
 & TSMixer       & \xmark & ---    & $8.66\times10^{-3}$ & 2.50 & $2.34\times10^{-4}$ & 15.88 & $6.06$  \\
 &               & \cmark & \textbf{0.0005} & \bm{$5.86\times10^{-3}$} & \textbf{1.15} & \bm{$6.35\times10^{-5}$} & \textbf{6.06}  & \bm{$0.092$} \\
\hline
\multirow{8}{*}{E}
 & TimeMixer     & \xmark & ---    & $6.63\times10^{-3}$ & 1.74 & $7.29\times10^{-5}$ & 7.95  & $9.85$  \\
 &               & \cmark & \textbf{0.0001} & \bm{$6.17\times10^{-3}$} & \textbf{0.82} & \bm{$6.50\times10^{-5}$} & \textbf{4.49}  & \bm{$6.76$}  \\
 & ITransformer  & \xmark & ---    & $7.15\times10^{-3}$ & 1.31 & $8.57\times10^{-5}$ & 5.40  & $17.00$ \\
 &               & \cmark & \textbf{0.001}  & \bm{$6.38\times10^{-3}$} & \textbf{1.31} & \bm{$6.89\times10^{-5}$} & \textbf{5.40}  & \bm{$7.69$}  \\
 & Mamba         & \xmark & ---    & $6.74\times10^{-3}$ & 1.71 & $7.74\times10^{-5}$ & 8.29  & $20.40$ \\
 &               & \cmark & \textbf{0.0001} & \bm{$6.19\times10^{-3}$} & \textbf{0.91} & \bm{$6.54\times10^{-5}$} & \textbf{4.56}  & \bm{$8.52$}  \\
 & TSMixer       & \xmark & ---    & *                  &  *   & *                  &  *   & *       \\
 &               & \cmark & \textbf{0.001}  & \bm{$6.20\times10^{-3}$} & \textbf{0.89} & \bm{$6.55\times10^{-5}$} & \textbf{4.59}  & \bm{$7.85$}  \\
\hline
\multirow{8}{*}{F}
 & TimeMixer     & \xmark & ---    & $7.32\times10^{-3}$ & 1.06 & $8.94\times10^{-5}$ & 4.59  & $2.62$  \\
 &               & \cmark & \textbf{0.001}  & \bm{$6.88\times10^{-3}$} & \textbf{0.73} & \bm{$8.08\times10^{-5}$} & \textbf{3.78}  & \bm{$2.31$}  \\
 & ITransformer  & \xmark & ---    & $7.53\times10^{-3}$ & 1.47 & $8.57\times10^{-5}$ & 6.80  & $6.33$  \\
 &               & \cmark & \textbf{0.0001} & \bm{$6.97\times10^{-3}$} & \textbf{0.78} & \bm{$8.29\times10^{-5}$} & \textbf{3.78}  & \bm{$2.71$}  \\
 & Mamba         & \xmark & ---    & $7.51\times10^{-3}$ & 1.64 & $9.62\times10^{-5}$ & 7.26  & $13.70$ \\
 &               & \cmark & \textbf{0.0005} & \bm{$6.95\times10^{-3}$} & \textbf{0.78} & \bm{$8.23\times10^{-5}$} & \textbf{3.78}  & \bm{$2.93$}  \\
 & TSMixer       & \xmark & ---    & $1.06\times10^{-2}$ & 1.06 & $4.59\times10^{-4}$ & 4.59  & $6.52$  \\
 &               & \cmark & \textbf{0.0001} & \bm{$6.88\times10^{-3}$} & \textbf{0.77} & \bm{$8.06\times10^{-5}$} & \textbf{3.76}  & \bm{$2.60$}  \\
\hline
\multirow{8}{*}{G}
 & TimeMixer     & \xmark & ---    & $5.87\times10^{-3}$ & 2.16 & $6.60\times10^{-5}$ & 14.54 & $15.40$ \\
 &               & \cmark & \textbf{0.001}  & \bm{$3.79\times10^{-3}$} & \textbf{1.18} & \bm{$3.23\times10^{-5}$} & \textbf{9.54}  & \bm{$11.40$} \\
 & ITransformer  & \xmark & ---    & $8.21\times10^{-3}$ & 1.94 & $1.11\times10^{-4}$ & 9.04  & $11.50$ \\
 &               & \cmark & \textbf{0.0005} & \bm{$7.14\times10^{-3}$} & \textbf{0.74} & \bm{$8.68\times10^{-5}$} & \textbf{3.78}  & \bm{$2.81$}  \\
 & Mamba         & \xmark & ---    & $1.13\times10^{-2}$ & 3.86 & $2.55\times10^{-3}$ & 72.10 & $32.60$ \\
 &               & \cmark & \textbf{0.0005} & \bm{$3.86\times10^{-3}$} & \textbf{1.22} & \bm{$3.30\times10^{-5}$} & \textbf{9.81}  & \bm{$12.10$} \\
 & TSMixer       & \xmark & ---    & $2.48\times10^{-2}$ & 7.17 & $1.20\times10^{-2}$ & 423.99 & $45.80$ \\
 &               & \cmark & \textbf{0.001}  & \bm{$3.79\times10^{-3}$} & \textbf{1.21} & \bm{$3.21\times10^{-5}$} & \textbf{9.65}  & \bm{$11.80$} \\
\hline
\end{tabular}
}
\label{tab:timeseries_baselines}
\end{table*}

\subsection{Analyzing Divergence in Autoregressive Forecasts}
\label{appendix:divergence_analysis}
We observe a divergence (See Fig.~\ref{fig:appendix_fail_case_with_update}) where predicted values grow unbounded in the predictions on Dataset E. The comparison involves the TSMixer baseline and our proposed PhysTimeMD variant. The divergence starts to happen around 250 timesteps across all three dimensions of the visualized atom. We hypothesize that the divergence stems from a mismatch between training and inference. During training, ground-truth inputs are used, while inference depends on the model’s own predictions, causing errors to accumulate over time. This distribution shift leads to unstable rollouts. PhysTimeMD reduces such instability through physics-informed constraints, resulting in more stable and physically plausible long-term predictions.

\subsection{Hyperparameters Setup}
The physics loss coefficient \(\lambda\) is selected from \(\{0.001, 0.0005, 0.0001\}\) through tuning, with the best-performing value reported. At each training step, \(M = 500\) atom pairs are sampled to compute the energy-based physics loss, which enforces physically plausible interactions. All backbone models are trained using a batch size \(B = 16\), an adaptive learning rate \(\eta = 0.01\), gradient clipping, and an early stopping strategy with a patience of 3 epochs, applied over a maximum of 10 epochs. This training setup ensures stable and efficient convergence while maintaining physical consistency in predictions.

\subsection{Benchmarking with Time-Series Baselines}
See Tab.~\ref{tab:timeseries_baselines} for detailed results of PhysTimeMD.